\pdfoutput=1

\documentclass[11pt]{article}

\usepackage[final]{acl}
\usepackage{times}
\usepackage{latexsym}
\usepackage{booktabs}
\usepackage{amsmath}
\usepackage{colortbl}
\usepackage{enumitem}
\usepackage{multirow}
\usepackage{xcolor}  
\usepackage{amssymb} 
\usepackage{forest}
\usepackage{array}
\usepackage{tikz}
\usepackage{hyperref}
\usepackage[utf8]{inputenc}
\usepackage{amssymb}
\usepackage{mathtools}

\newcommand{\greenTriangle}{\tikz\draw[scale=0.2, fill=green!70!black, rotate=90] (0,0) -- (1,0.5) -- (0,1) -- cycle;}
\newcommand{\redCircle}{\tikz\draw[scale=0.2, fill=red] (0,0) circle (0.5);}

\usepackage[T1]{fontenc}

\usepackage[utf8]{inputenc}

\usepackage{microtype}

\usepackage{inconsolata}

\usepackage{graphicx}

\title{A Survey of Pun Generation: Datasets, Evaluations and Methodologies}

\author{
 \textbf{Yuchen Su\textsuperscript{1}\thanks{Corresponding author}},
 \textbf{Yonghua Zhu\textsuperscript{2}},
 \textbf{Ruofan Wang\textsuperscript{1}},
 \textbf{Zijian Huang\textsuperscript{1}},
 \\
 \textbf{Diana Benavides-Prado \textsuperscript{3}},
 \textbf{Michael Witbrock\textsuperscript{1}},
\\
 \textsuperscript{1}School of Computer Science, University of Auckland, New Zealand
 \\
 \textsuperscript{2}Singapore University of Technology and Design, Singapore
 \\
 \textsuperscript{3}School of Electronic Engineering and Computer Science, Queen Mary University of London
 \\
 \texttt{\{ysu132, rwan551, zhua764\}@aucklanduni.ac.nz, yonghua\_zhu@sutd.edu.sg}
 \\
 \texttt{d.benavidesprado@qmul.ac.uk, m.witbrock@auckland.ac.nz}
}

\begin{document}
\maketitle
\begin{abstract}
Pun generation seeks to creatively modify linguistic elements in text to produce humour or evoke double meanings. It also aims to preserve coherence and contextual appropriateness, making it useful in creative writing and entertainment across various media and contexts. Although pun generation has received considerable attention in computational linguistics, there is currently no dedicated survey that systematically reviews this specific area. To bridge this gap, this paper provides a comprehensive review of pun generation datasets and methods across different stages, including conventional approaches, deep learning techniques, and pre-trained language models. Additionally, we summarise both automated and human evaluation metrics used to assess the quality of pun generation. Finally, we discuss the research challenges and propose promising directions for future work. \footnote{https://github.com/ysu132/Pun-Generation-Survey}
\end{abstract}

\section{Introduction}
A pun is a kind of rhetorical style that leverages the polysemy or phonetic similarity of words to produce expressions with double or multiple meanings \cite{delabastita2016traductio}. Beyond mere wordplay, puns serve as a crucial mechanism of linguistic creativity, enriching communication and making it more engaging \cite{carter2015language}. For example, the pun sentence ``I used to be a banker, but I lost interest'' plays on the pun words ``interest'', encompassing both a lack of enthusiasm for banking as a profession and the idea of financial loss. This ability to encode multiple layers of meaning fosters cognitive flexibility, encouraging individuals to interpret language in innovative ways  \cite{zheng2023humor}. Due to the unique capacity of puns, they are widely used in advertising \cite{djafarova2008advertisers,van2005puns}, literature \cite{giorgadze2014linguistic}, and various other fields.

%
Natural language generation (NLG) tasks involve the creation of human-like text by computers based on given data or input \cite{gatt2018survey}, with pun generation being a notable and challenging aspect of such tasks. There are various approaches utilised in automatic pun generation, including template-based methods \cite{hong2009automatically}, deep neural network approaches \cite{he2019pun}, and pre-trained language models (PLMs) employing various training and inference styles \cite{mittal2022ambipun,xu2024good}. These methods are applied to different types of puns, with a particular focus on homophonic \cite{yu-etal-2020-homophonic}, homographic \cite{yu2018neural,luo2019pun}, heterographic puns \cite{xu2024good} and visual puns \cite{rebrii2022words}.

Despite the long-standing research interest in pun generation, a comprehensive literature review in this field has not been conducted, to the best of our knowledge. Some existing relevant surveys focus on generating creative writing and explore tasks such as poetry composition \cite{bena2020introducing,elzohbi2023creative}, storytelling \cite{gieseke2021survey,alhussain2021automatic}, arts \cite{shahriar2022gan} and metaphor \cite{rai2020survey,ge2023survey}. It is noteworthy that  \citet{amin2020survey} outlined methods to humour generation, discussing various systems based on templates and neural networks, along with their respective strengths and weaknesses. However, they did not cover the pun research nor incorporate relevant technologies associated with large language models (LLMs). Therefore, we aim to address this gap by conducting the first comprehensive survey on pun generation, which can provide valuable guidance for researchers engaged in the study of puns.

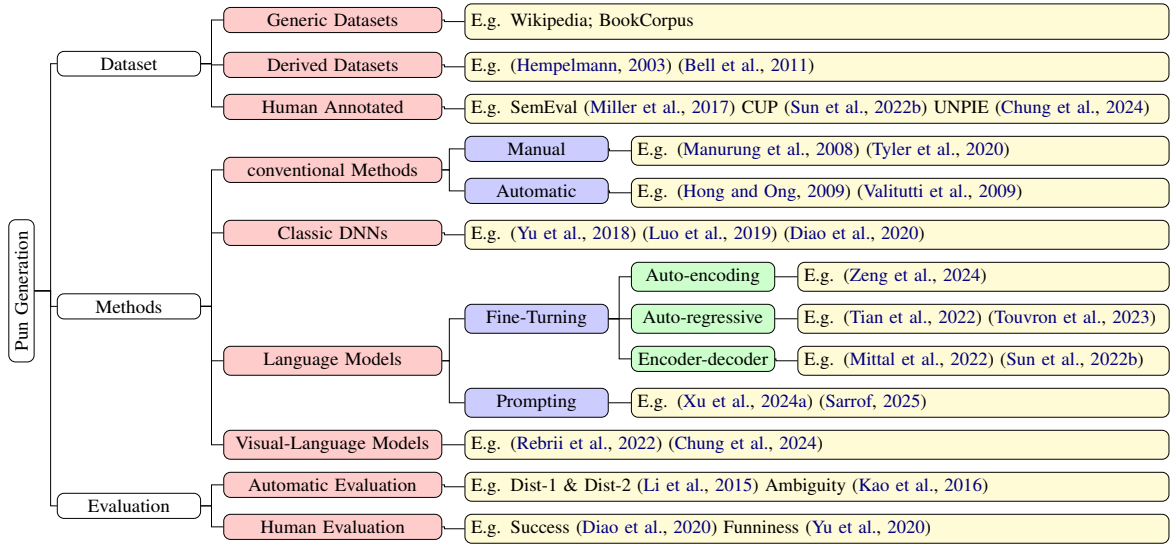
\begin{figure*}
    \centering
    \resizebox{2\columnwidth}{!}{
    \begin{forest}
        for tree={grow'=east, draw, rounded corners, text width=7em, 
                  edge path={\noexpand\path[\forestoption{edge}] (!u.parent anchor)
                      -- +(0.6cm,0) |- (.child anchor)\forestoption{edge label};},
                  fit=band}
        [Pun Generation,
            rotate=90,
            node options={align=center},
            [Dataset,
                node options={align=center},
                [Generic Datasets,
                    node options={align=center},
                    text width=11em, 
                    fill=red!20, 
                    edge path={\noexpand\path[\forestoption{edge}] (!u.parent anchor)
                      -- +(1.7cm,0) |- (.child anchor)\forestoption{edge label};}
                    [
                    E.g. Wikipedia; BookCorpus,
                    align=left, 
                    fill=yellow!20,
                    text width=37em,
                        edge path={\noexpand\path[\forestoption{edge}] (!u.parent anchor)
                      -- +(2.3cm,0) |- (.child anchor)\forestoption{edge label};}
                    ]
                ]
                [Derived Datasets,
                    node options={align=center},
                    text width=11em, 
                    fill=red!20,
                    edge path={\noexpand\path[\forestoption{edge}] (!u.parent anchor)
                      -- +(1.7cm,0) |- (.child anchor)\forestoption{edge label};}
                    [
                    E.g. \cite{hempelmann2003paronomasic} \cite{bell2011wordplay},
                    fill=yellow!20,
                    text width=37em,
                        edge path={\noexpand\path[\forestoption{edge}] (!u.parent anchor)
                      -- +(2.3cm,0) |- (.child anchor)\forestoption{edge label};}
                    ]
                ]
                [Human Annotated,
                    node options={align=center},
                    text width=11em, 
                    fill=red!20,
                    edge path={\noexpand\path[\forestoption{edge}] (!u.parent anchor)
                      -- +(1.7cm,0) |- (.child anchor)\forestoption{edge label};}
                    [
                    E.g. SemEval \cite{miller2017semeval} CUP \cite{sun2022context} UNPIE \cite{chung2024can},
                    fill=yellow!20,
                    text width=37em,
                        edge path={\noexpand\path[\forestoption{edge}] (!u.parent anchor)
                      -- +(2.3cm,0) |- (.child anchor)\forestoption{edge label};}
                    ]
                ]
            ]
            [Methods,
                node options={align=center},
                [conventional Methods,
                    text width=11em, 
                    node options={align=center},
                    fill=red!20,
                    edge path={\noexpand\path[\forestoption{edge}] (!u.parent anchor)
                      -- +(1.7cm,0) |- (.child anchor)\forestoption{edge label};}
                    [
                    Manual,
                    fill=blue!20,
                        node options={align=center},
                        edge path={\noexpand\path[\forestoption{edge}] (!u.parent anchor)
                      -- +(2.4cm,0) |- (.child anchor)\forestoption{edge label};}
                      [
                        E.g. \cite{manurung2008construction} \cite{tyler2020computational},
                        fill=yellow!20,
                        text width=28.1em,
                        edge path={\noexpand\path[\forestoption{edge}] (!u.parent anchor)
                      -- +(1.5cm,0) |- (.child anchor)\forestoption{edge label};}
                      ]
                    ]
                    [
                    Automatic,
                    fill=blue!20,
                    node options={align=center},
                        edge path={\noexpand\path[\forestoption{edge}] (!u.parent anchor)
                      -- +(2.4cm,0) |- (.child anchor)\forestoption{edge label};}
                      [
                        E.g. \cite{hong2009automatically} \cite{valitutti2009graphlaugh},
                        text width=28.1em,
                        fill=yellow!20,
                        edge path={\noexpand\path[\forestoption{edge}] (!u.parent anchor)
                      -- +(1.5cm,0) |- (.child anchor)\forestoption{edge label};}
                      ]
                    ]
                ]
                [Classic DNNs,
                    text width=11em, 
                    node options={align=center},
                    fill=red!20,
                    edge path={\noexpand\path[\forestoption{edge}] (!u.parent anchor)
                      -- +(1.7cm,0) |- (.child anchor)\forestoption{edge label};}
                    [
                      E.g. \cite{yu2018neural} \cite{luo2019pun} \cite{diao2020afpun},
                        text width=37em,
                        fill=yellow!20,
                        edge path={\noexpand\path[\forestoption{edge}] (!u.parent anchor)
                      -- +(2.3cm,0) |- (.child anchor)\forestoption{edge label};}
                    ]
                ]
                [Language Models,
                    text width=11em, 
                    node options={align=center},
                    fill=red!20,
                    edge path={\noexpand\path[\forestoption{edge}] (!u.parent anchor)
                      -- +(1.7cm,0) |- (.child anchor)\forestoption{edge label};}
                    [
                    Fine-Turning,
                    fill=blue!20,
                    node options={align=center},
                        edge path={\noexpand\path[\forestoption{edge}] (!u.parent anchor)
                      -- +(2.4cm,0) |- (.child anchor)\forestoption{edge label};}
                      [
                        Auto-encoding,
                        fill=green!20,
                        node options={align=center},
                        edge path={\noexpand\path[\forestoption{edge}] (!u.parent anchor)
              -- +(1.7cm,0) |- (.child anchor)\forestoption{edge label};}
                            [
                                E.g. \cite{zeng2024barking},
                                text width=19.2em,
                                fill=yellow!20,
                                edge path={\noexpand\path[\forestoption{edge}] (!u.parent anchor)
                              -- +(1.6cm,0) |- (.child anchor)\forestoption{edge label};}
                              ]
                      ]
                      [
                      Auto-regressive,
                      fill=green!20,
                        node options={align=center},
                        edge path={\noexpand\path[\forestoption{edge}] (!u.parent anchor)
              -- +(1.7cm,0) |- (.child anchor)\forestoption{edge label};}
                            [
                                E.g. \cite{tian2022unified} \cite{touvron2023llama},
                                text width=19.2em,
                                fill=yellow!20,
                                edge path={\noexpand\path[\forestoption{edge}] (!u.parent anchor)
                              -- +(1.6cm,0) |- (.child anchor)\forestoption{edge label};}
                              ]
                      ]
                      [
                      Encoder-decoder,
                      fill=green!20,
                        node options={align=center},
                        edge path={\noexpand\path[\forestoption{edge}] (!u.parent anchor)
                  -- +(1.7cm,0) |- (.child anchor)\forestoption{edge label};}
                            [
                                E.g. \cite{mittal2022ambipun} \cite{sun2022context},
                                text width=19.2em,
                                fill=yellow!20,
                                edge path={\noexpand\path[\forestoption{edge}] (!u.parent anchor)
                              -- +(1.5cm,0) |- (.child anchor)\forestoption{edge label};}
                              ]
                      ]
                    ]
                    [
                    Prompting,
                    fill=blue!20,
                    node options={align=center},
                        edge path={\noexpand\path[\forestoption{edge}] (!u.parent anchor)
                      -- +(2.4cm,0) |- (.child anchor)\forestoption{edge label};}
                      [
                        E.g. \cite{xu2024good} \cite{sarrof2025homophonic},
                        text width=28.1em,
                        fill=yellow!20,
                        edge path={\noexpand\path[\forestoption{edge}] (!u.parent anchor)
                      -- +(1.6cm,0) |- (.child anchor)\forestoption{edge label};}
                      ]
]
                ]
                [
                Visual-Language Models,
                    text width=11em, 
                    node options={align=center},
                    fill=red!20,
                    edge path={\noexpand\path[\forestoption{edge}] (!u.parent anchor)
                      -- +(1.7cm,0) |- (.child anchor)\forestoption{edge label};}
                    [
                      E.g. \cite{rebrii2022words} \cite{chung2024can},
                        text width=37em,
                        fill=yellow!20,
                        edge path={\noexpand\path[\forestoption{edge}] (!u.parent anchor)
                      -- +(2.4cm,0) |- (.child anchor)\forestoption{edge label};}
                    ]
                ]
            ]
            [Evaluation,
            node options={align=center},
                [
                Automatic Evaluation,
                    text width=11em, 
                    node options={align=center},
                    fill=red!20,
                    edge path={\noexpand\path[\forestoption{edge}] (!u.parent anchor)
                      -- +(1.7cm,0) |- (.child anchor)\forestoption{edge label};}
                    [
                      E.g. Dist-1 \& Dist-2 \cite{li2015diversity} Ambiguity \cite{kao2016computational},
                      fill=yellow!20,
                        text width=37em,
                        edge path={\noexpand\path[\forestoption{edge}] (!u.parent anchor)
                      -- +(2.3cm,0) |- (.child anchor)\forestoption{edge label};}
                      ]
                ]
                [
                Human Evaluation,
                    text width=11em, 
                    node options={align=center},
                    fill=red!20,
                    edge path={\noexpand\path[\forestoption{edge}] (!u.parent anchor)
                      -- +(1.7cm,0) |- (.child anchor)\forestoption{edge label};}
                    [
                      E.g. Success \cite{diao2020afpun} Funniness \cite{yu-etal-2020-homophonic},
                      fill=yellow!20,
                      text width=37em,
                        edge path={\noexpand\path[\forestoption{edge}] (!u.parent anchor)
                      -- +(2.3cm,0) |- (.child anchor)\forestoption{edge label};}
                      ]
                ]
            ]
        ]
        \end{forest}
    }
    \caption{The survey tree for pun generation.}
    \label{survey tree}
\end{figure*}

In this survey, we review the past three decades of research and examine the current state of natural language pun generation, analysing the datasets and categorising these methods in five groups based on their technological development timeline: (1) Conventional methods, which involve generating puns by manually or automatically constructing templates; (2) Classic Deep Neural networks (DNNs), leveraging architectures, such as RNNs and their variants, to learn pun patterns from data; (3) Fine-tuning of PLMs, where pre-trained models like GPT \cite{radford2018improving} are adapted with task-specific datasets to improve pun generation, (4) Prompting of PLMs, which utilizes carefully designed prompts to guide models in generating puns without additional training, and (5) Visual-language models, where some preliminary studies on visual pun generation. We further summarise the automatic and human evaluation metrics used to assess the quality of generated puns. Finally, we discuss our findings and propose promising research directions for future work in this field. 

Overall, the paper is organised as follows: Section \ref{task description} reviews the main categories of puns and provides examples for each category. Section \ref{pundata}, \ref{Methodology} and \ref{evaluation} summarise the relevant datasets, methods, and evaluation metrics, as shown in figure \ref{survey tree}. We also discuss the challenges and outline future research directions in Section \ref{future}, as well as conclude with final remarks in Section \ref{conclusion}.



\section{Pun Categories}
\label{task description}
This section outlines the main four types of puns: i) \textit{Homophonic puns}, ii) \textit{Heterographic puns}, iii) \textit{Homographic puns} and iv) \textit{Visual pun}. 
\subsection{Homophonic Puns}
Homophonic puns rely on the dual meanings of homophones, which are words that sound alike but have different meanings \cite{attardo2009linguistic}. This is illustrated in example \ref{example1}:
\begin{enumerate}[label=(\alph*)]
    \item \label{example1} Dentists don't like a hard day at the \underline{orifice} (office).
\end{enumerate}
which uses the ``orifice'' as the pivotal pun word. The term ``orifice'' refers to the human mouth, while its pronunciation is similar to ``office''. This similarity allows it to be interpreted as a dentist working in an office, thereby creating a humorous pun effect.

\subsection{Heterographic Puns}
Heterographic puns emphasise differences in spelling with the same pronunciation to achieve their rhetorical effect, which are also classified as homophonic puns in some studies \cite{sun2022context,miller2017semeval}. An example of a heterographic pun is shown as \ref{example2}: 
\begin{enumerate}[label=(\alph*)]
\setcounter{enumi}{1}
    \item \label{example2} Life is a puzzle, look here for the missing \underline{peace} (piece). \cite{xu2024good}
\end{enumerate}
The word "peace" can be interpreted as tranquility in life, while it shares the same pronunciation as "piece" which refers to a puzzle piece. Therefore, the pun can be recognized as seeking either peace in life or the missing piece of a puzzle. 

\subsection{Homographic Puns}
Homographic puns exploit words spelled the same homographs but possess different meanings \cite{attardo2009linguistic}, as shown in example \ref{example3}:
\begin{enumerate}[label=(\alph*)]
\setcounter{enumi}{2}
    \item \label{example3} Always trust a glue salesman. They tend to \underline{stick} to their word.
\end{enumerate}
The phrase ``stick to their word'' refers to the act of keeping a promise in common English expressions. However, the meaning of ``stick'' is also directly associated with the adhesive properties of ``glue'', which artfully plays on the dual meanings of the word ``stick''.

\subsection{Visual Puns}
\begin{figure}[t]
\centering
      \resizebox{0.7\columnwidth}{!}{
  \includegraphics[width=\columnwidth,height=0.6\columnwidth]{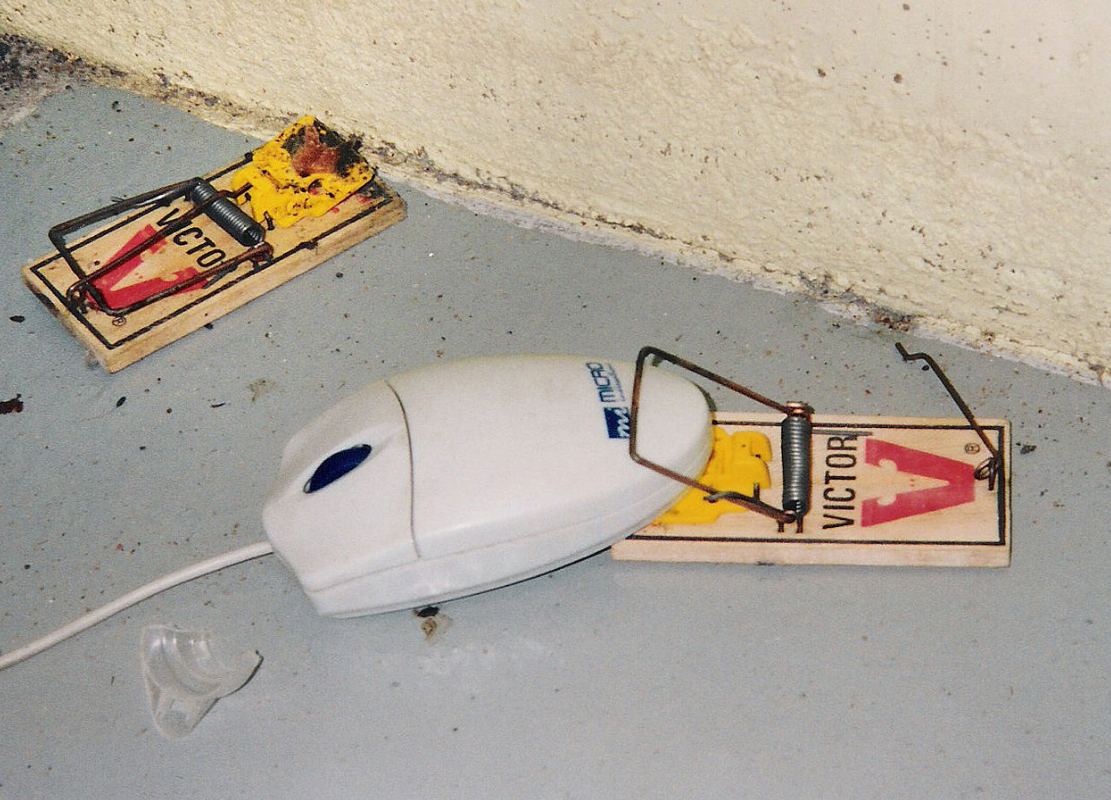}
  }
  \caption{A visual pun example features a white mouse and a mousetrap, where the combination exploits the double meaning of the word ``mouse''.}
  \label{visual pun}
\end{figure}

Visual puns are a form of artistic expression that utilises images or visual elements to create double meanings \cite{smith2008impact}. A typical example of a visual pun from Wikipedia \footnote{https://en.wikipedia.org/wiki/Visual\_pun} is shown in Figure \ref{visual pun}. The figure leverages the multiple meanings of the word "mouse" based on the computer device and animal, thereby creating a pun effect by combining the computer mouse and mousetrap.

\section{Dataset}
\label{pundata}
In this section, we present the current datasets that have been used and constructed for pun research. We classified the datasets into generic datasets, derived datasets and human-annotated datasets. For the detailed table of the pun dataset, please refer to Appendix \ref{pundataset}.
\subsection{Generic Datasets}
In the early days of neural network technology, due to the difficulty in obtaining adequate data to train seq2seq models for some specific tasks \cite{yu2018neural}, most research in pun generation relied on general datasets to train conditional language models, enabling them to capture fundamental semantic relationships. For example, some pun generation studies use the English Wikipedia corpus to train the language model \cite{yu2018neural,luo2019pun,diao2020afpun}, while others rely on BookCorpus \cite{zhu2015aligning,yu-etal-2020-homophonic} as a generic corpus for retrieval and training. \citet{sarrof2025homophonic} proposed a corss-lingual homophone identification algorithm and analysed the distribution of Hindi words in Latin and Devanagari scripts using C4 \cite{raffel2020exploring} and The Pile \cite{gao2020pile}, and then tested on the Dakshina dataset \cite{roark2020processing}.

\subsection{Derived Datasets}

The derived datasets are created as the new datasets by processing, transforming, or extracting specific details from general data. In this section, we present a list of derived datasets and outline the domains used in their creation. \citet{sobkowiak1991metaphonology} collected 3850 puns from advertisements and conversation, while \citet{hempelmann2003paronomasic} selected a subset for the automatic generation of heterophonic puns. \citet{lucas2004deciphering} proposed a tiny pun corpus that relies on lexical ambiguity from newspaper comics. \citet{bell2011wordplay} created a 373 puns dataset from church marquees and literature to study wordplay in religious advertising. In addition, several studies have created pun datasets by filtering data from specialised joke websites. For example, both \citet{yang2015humor} and \citet{kao2016computational} curated pun datasets by crawling data from the "Pun of the Day" website. \citet{jaech2016phonological} compiled a homophonic pun dataset from Tumblr, Reddit, and Twitter to facilitate the automatic recovery of the target word in given puns.

\subsection{Human Annotated}
This section provides some details of human-annotated pun datasets. \textbf{SemEval.} \citet{miller2017semeval} released two manually annotated pun datasets based on \cite{miller2016towards} and \cite{miller2016adjusting} including both homophonic and heterographic puns, which is one of the most commonly used datasets in the pun generation community. \textbf{SemEval Enhancements.} \citet{sun2022context} augmented the SemEval dataset by adding pun data combined with a given context and provided annotations on the adaptation between context words and their corresponding pun pairs. Furthermore, \citet{sun2022expunations} added the fine-grained funniness ratings and natural language explanations based on the SemEval dataset. \textbf{ChinesePun.} \citet{chen2024u} introduced the first datasets for Chinese homophonic and homographic puns, specifically designed for pun understanding and generation tasks. \textbf{Multimodal Dataset.} \citet{zhang2024creating} compiled a large collection of Chinese historical visual puns and provided detailed annotations, including the identification of prominent visual elements, matching of these elements with their symbolic meanings and interpretations. \citet{chung2024can} selected a subset of homophonic and heterogeneous puns from the SemEval dataset and supplemented it with corresponding explanation images.

\section{Methodology}
In this section, we provide an overview of existing approaches to pun generation.
\label{Methodology}

\subsection{Conventional Models}
\label{Traditional Models}
Early conventional methods are typically through template-based construction. In linguistics, a template refers to a textual structure consisting of predefined slots that can be populated with various variables \cite{amin2020survey}. \citet{binsted1994implemented} developed the simple question-answer system of pun-generator Joke Analysis and Production Engine (JAPE), which was improved in subsequent versions including JAPE-2 \cite{binsted1996machine} and JAPE-3. The model incorporates two primary structures: schemata, which are used to explore the relationships between different keywords, and templates, which are designed to generate the basic framework for puns. Inspired by JAPE, \citet{manurung2008construction} designed the STANDUP system, which expands and varies the elements generated by puns through further semantic and phonological analysis, for children with complex communication needs. Furthermore, \citet{tyler2020computational} expanded upon the JAPE system by incorporating more recent knowledge bases and designed the PAUL BOT system, enhancing its capabilities and flexibility in automated pun generation. 

Additionally, HCPP \cite{venour2000computational} and WISCRAIC \cite{mckay2002generation} systems both implement models for the specific subclass of puns about homonym common phrase and idiom-based witticisms according to semantic associations, respectively. \citet{hempelmann2003paronomasic} studies target recoverability, arguing that a robust model for target alternative words recovery provides the necessary foundation for heterographic pun generation.  \citet{ritchie2005computational} considered pun generation from the broader perspective of NLG. They analyse the differences in mechanisms between pun generation and conventional NLG, as well as the computational methods that could potentially accomplish this task. As for the research on non-English puns, \citet{dybala2008humor} designed a Japanese pun generator as part of a conversational system, while \citet{dehouck2025rule} proposed a generator for automatically generating French puns based on a given name and a word or phrase using rules.

Since building templates manually is a tedious and time-consuming task, \citet{hong2009automatically} proposed Template-Based
Pun Extractor and Generator (T-PEG) automatically identify, extract and represent the word relationships in a template, and then use these templates as patterns for the computer to generate its own puns.  \citet{valitutti2009graphlaugh} generated funny puns by implementing GraphLaugh to automatically generate different types of lexical associations and visualize them through a dynamic graph. They also explored a method for automatically generating humour through the substitution of words in short texts \cite{valitutti2013let}.



\begin{table*}
  \centering
  \resizebox{2\columnwidth}{!}{
  \begin{tabular}{lcccc}
    \toprule
    \textbf{Method} & \textbf{Model} & \textbf{Type} & \textbf{Language} & \textbf{Dataset}   \\
    \midrule
    \rowcolor{gray!20}\multicolumn{5}{c}{Classic Deep Neural Networks }\\
    Neural Pun \cite{yu2018neural} & LSTM  &hog&English&Wikipedia \& \cite{miller2017semeval}\\
    Pun-GAN \cite{luo2019pun} & LSTM &hog&English&Wikipedia \& \cite{miller2017semeval}\\
    SurGen \cite{he2019pun} & LSTM &hop&English&BookCorpus \& \cite{miller2017semeval}\\
    LCR \cite{yu-etal-2020-homophonic} & LSTM &hop&English& BookCorpus \& \cite{hu2019parabank} \\
    AFPun-GAN \cite{diao2020afpun} & ON-LSTM &hog&English& Wikipedia \& \cite{miller2017semeval} \\
    \midrule
    \rowcolor{gray!20}\multicolumn{5}{c}{Pre-trained Language Models}\\
    Ext Ambipun\cite{mittal2022ambipun}     &  T5 &hog&English&\cite{annamoradnejad2020colbert}          \\
    Sim Ambipun\cite{mittal2022ambipun}     &  T5     &hog&English&\cite{annamoradnejad2020colbert}      \\
    Gen Ambipun\cite{mittal2022ambipun}     &  T5    &hog&English&\cite{annamoradnejad2020colbert}       \\
    UnifiedPun\cite{tian2022unified}     & GPT-2 \& BERT     &hog\&hog&English&\cite{annamoradnejad2020colbert}       \\
    Context-pun\cite{sun2022context} & T5 &hog\&heg&English&\cite{sun2022context} \\
    PunIntended \cite{zeng2024barking} & BERT  &hop\&hog&English&\cite{sun2022expunations}\\
    PGCL \cite{chen2024u}  & LLaMA2-7B  &hop\&hog&English&\cite{miller2017semeval}  \\
    PGCL \cite{chen2024u}  & Baichuan2-7B  &hop\&hog&Chinese&\cite{chen2024u}  \\
     Hinglish \cite{sarrof2025homophonic} &GPT-3.5&hop& Multi-language & C4 \& The Pile \& Dakshina\\
    \bottomrule
  \end{tabular}
  }
  \caption{Methods of neural network models and pre-trained language models for pun generation task. Hog, hop and heg denote the types of homographic puns, homophonic puns and heterographic puns, respectively.}
  \label{methods1}
\end{table*}

\subsection{Classic DNNs}
With the development of deep learning, pun generation has increasingly been implemented using deep neural networks, including Sequence-to-Sequence (Seq2Seq) \cite{sutskever2014sequence} and Generative Adversarial Network (GAN) \cite{goodfellow2014generative}. In general, Seq2Seq models map input sequences, such as words and phrases, to output the pun sentence, by maximising the conditional log-likelihood of the generated sequence.

\citet{yu2018neural} represented the first attempt to apply deep neural networks to generate homographic puns without specific training data by developing a conditional language model \cite{Mou2015BackwardAF} that creates sentences containing a target word with dual meanings. Building on this generator, \citet{luo2019pun} introduced a novel discriminator, which is a word sense classifier with a single-layer bi-directional LSTM, to provide a well-structured ambiguity reward for the generator. \citet{diao2020afpun} replaced the conventional LSTM network structure with ON-LSTM \cite{shen2018ordered} to further enhance performance. Additionally, \citet{he2019pun} and \citet{yu-etal-2020-homophonic} used the Seq2Seq model to rewrite the sentence so that it remains grammatically correct after replacing pun words. 

In general, classic DNNs can generate puns that are more flexible compared to conventional models by fitting both general and pun datasets. However, existing methods heavily rely on annotated data and limited types of corpora, which restricts further improvement in the quality of pun generation.


\subsection{Pre-trained Language Models}
Early PLMs, such as Word2Vec \cite{mikolov2013efficient} and GloVe \cite{pennington2014glove}, are distributed word representation methods trained on large-scale unlabeled text data, capable of capturing both the semantic and contextual information of words. These models are utilised to address various sub-tasks involved in pun generation, which has a bunch of semantic prior knowledge than classic DNNs. For example, \citet{mittal2022ambipun} proposed to get the context words from Word2Vec based on pun words. \citet{yu-etal-2020-homophonic} designed a constraint selection algorithm based on lexical semantic relevance and obtained the word embeddings from Continuous Bag of Words (CBOW) \cite{mikolov2013efficient}.

Most contemporary PLMs are built upon the Transformer architecture \cite{vaswani2017attention}, which has shown outstanding performance across various natural language processing tasks \cite{min2023recent}. The main model categories are classified into: (1) auto-encoding models, such as BERT \cite{devlin-etal-2019-bert}, (2) auto-regressive models, such as the GPT-2 \cite{radford2019language}, and (3) encoder-decoder models, such as T5 \cite{raffel2020exploring}. Pun generation tasks are primarily implemented through fine-tuning and prompting strategies.

\subsubsection{PLMs with Fine-Tuning}
Fine-tuning PLMs is to further train the model on a specific dataset to make it better suited to the needs of a specific task. For auto-encoding models, since the bidirectional encoding characteristics of the model are not suitable for generation tasks, most current work on pun generation employs it as the discriminator in GANs. For example, \citet{zeng2024barking} and \citet{tian2022unified} both used the BERT-base model, leveraging the [CLS] token representation for classification. 

In auto-regressive models, \citet{tian2022unified} finetuned the GPT-2 model based on the combination dataset of Gutenberg BookCorpus and jokes \cite{annamoradnejad2020colbert} and proposed a unified framework for generating both homophonic and homographic puns. \citet{chen2024u} finetuned both LLaMA2-7B \cite{touvron2023llama} and Baichuan2-7B \cite{yang2023baichuan} for generating English and Chinese puns respectively through the standard Direct Preference Optimization \cite{rafailov2024direct} and multistage curriculum learning framework.

For encoder-decoder models, \citet{mittal2022ambipun} explored the generation of puns based on context words associated with pun words and finetuned a keyword-to-sentence model using the T5 model. Similarly, \citet{sun2022context} proposed the context-situated pun generation, which involves identifying pun words for a given set of contextual keywords and then generating puns based on these keywords and the associated pun words. \citet{zeng2024barking} used T5 as a generator, taking the pun semantic trees as input and generating pun text as output.

\subsubsection{PLMs with Prompting}
Prompting \cite{Liu2021PretrainPA} refers to a specially designed input mode intended to guide PLMs, especially for LLMs, in performing specific tasks \cite{Alhazmi2024DistractorGI}. However, there are few studies exploring pun generation specifically from the perspective of prompting. \citet{mittal2022ambipun} provides examples of the target pun along with its two interpretations and instructions for generating the pun in GPT-3 \cite{Brown2020LanguageMA} to serve as a baseline comparison model. Based on the Chain-of-Thought prompting approach \cite{wei2022chain}, \citet{sarrof2025homophonic} designed a novel method that integrates homophone and transliteration modules to enhance the quality of pun generation. 

In addition, \citet{xu2024good} selected a range of prominent LLMs to evaluate their capabilities on pun generation, including both open-source models in Llama2-7B-Chat \cite{touvron2023llama}, Mistral-7B \cite{Jiang2023Mistral7}, Vicuna-7B \cite{Zheng2023JudgingLW}, and OpenChat-7B \cite{Wang2023OpenChatAO}, and closed-source models in Gemini-Pro \cite{2023arXiv231211805G}, GPT-3.5-Turbo \cite{openai2023gpt35}, Claude3-Opus \cite{TheC3}, and GPT-4-Turbo \cite{openai2023gpt4}. These studies reveal that although LLMs still exhibit limitations in generating creative and humorous puns, their demonstrated potential highlights a developmental trend in this field. Future research can further optimize existing LLMs to enhance their performance in pun generation tasks.

\subsection{Visual-Language Models}
\label{visual-language}
There are currently some preliminary studies on visual puns. \citet{rebrii2022words} explored the cross-lingual translation of puns combined with visual elements. \citet{chung2024can} employed the DALLE-3 \cite{betker2023improving} to generate images that illustrated the meanings of puns based on textual puns. \citet{zhang2024creating} leveraged their established dataset to conduct a comprehensive evaluation of large vision-language models in visual pun comprehension. However, to the best of our knowledge, there are no dedicated studies on visual pun generation, which is a potential research direction.

\section{Evaluation Strategies}
\label{evaluation}

\begin{table*}
    \centering
    \resizebox{2\columnwidth}{!}{
    \begin{tabular}{l|ccccccc|cccccc}
    \toprule
    \multirow{2}{*}{Paper} & \multicolumn{7}{c|}{Automatic Evaluation} & \multicolumn{6}{c}{Human Evaluation}\\
    
    &  PPLs. & D1\&2. & Succ. & Ambi. & Dist & Surp. & Unus.&Succ.&Funn. & Flun. & Info. & Cohe. & Read.\\
    \midrule
    \cite{yu2018neural}  & \textcolor{green}{\checkmark} &  \textcolor{green}{\checkmark} & \textcolor{red}{$\times$} & \textcolor{red}{$\times$} &\textcolor{red}{$\times$} &\textcolor{red}{$\times$} & \textcolor{red}{$\times$} & \textcolor{red}{$\times$} &  \textcolor{red}{$\times$} & \textcolor{green}{\checkmark}& \textcolor{red}{$\times$}&  \textcolor{green}{\checkmark}&  \textcolor{green}{\checkmark}\\
    
     \cite{he2019pun} & \textcolor{red}{$\times$} & \textcolor{red}{$\times$}& \textcolor{red}{$\times$} &  \textcolor{green}{\checkmark} &  \textcolor{green}{\checkmark} & \textcolor{green}{\checkmark} & \textcolor{green}{\checkmark}  & \textcolor{green}{\checkmark} &  \textcolor{green}{\checkmark} & \textcolor{red}{$\times$}& \textcolor{red}{$\times$}&  \textcolor{red}{$\times$}&  \textcolor{red}{$\times$}\\
     
     \cite{luo2019pun} &\textcolor{red}{$\times$} &\textcolor{green}{\checkmark}& \textcolor{red}{$\times$} &\textcolor{red}{$\times$}  &\textcolor{red}{$\times$} &\textcolor{red}{$\times$} & \textcolor{green}{\checkmark} & \textcolor{green}{\checkmark} &  \textcolor{red}{$\times$} & \textcolor{green}{\checkmark} & \textcolor{red}{$\times$}&  \textcolor{red}{$\times$}&  \textcolor{red}{$\times$}\\
     
      \cite{yu-etal-2020-homophonic} &\textcolor{red}{$\times$} &\textcolor{green}{\checkmark}& \textcolor{red}{$\times$} &\textcolor{red}{$\times$}  &\textcolor{red}{$\times$} &\textcolor{red}{$\times$} &\textcolor{red}{$\times$}  & \textcolor{green}{\checkmark} &  \textcolor{green}{\checkmark} & \textcolor{green}{\checkmark} & \textcolor{red}{$\times$}&  \textcolor{red}{$\times$}&  \textcolor{red}{$\times$}\\
      
      \cite{diao2020afpun} & \textcolor{red}{$\times$} & \textcolor{green}{\checkmark}& \textcolor{red}{$\times$} &\textcolor{red}{$\times$}  &\textcolor{red}{$\times$} &\textcolor{red}{$\times$} & \textcolor{green}{\checkmark} & \textcolor{green}{\checkmark} &  \textcolor{red}{$\times$} & \textcolor{green}{\checkmark} & \textcolor{red}{$\times$}&  \textcolor{red}{$\times$}&  \textcolor{red}{$\times$}\\
      
      \cite{mittal2022ambipun} &\textcolor{red}{$\times$} &\textcolor{green}{\checkmark}& \textcolor{red}{$\times$} &\textcolor{red}{$\times$}  &\textcolor{red}{$\times$} &\textcolor{red}{$\times$} &\textcolor{red}{$\times$}  & \textcolor{green}{\checkmark} &  \textcolor{green}{\checkmark} & \textcolor{red}{$\times$}  &  \textcolor{red}{$\times$} & \textcolor{green}{\checkmark}&  \textcolor{red}{$\times$}\\
      
      \cite{tian2022unified} & \textcolor{red}{$\times$}& \textcolor{red}{$\times$}& \textcolor{red}{$\times$} & \textcolor{green}{\checkmark} &\textcolor{green}{\checkmark} &\textcolor{green}{\checkmark} & \textcolor{red}{$\times$} & \textcolor{green}{\checkmark} &  \textcolor{green}{\checkmark} & \textcolor{red}{$\times$}  &  \textcolor{green}{\checkmark}&  \textcolor{red}{$\times$}&  \textcolor{red}{$\times$}\\

      \cite{sun2022context} &\textcolor{red}{$\times$} &\textcolor{red}{$\times$} & \textcolor{green}{\checkmark} &\textcolor{red}{$\times$}  &\textcolor{red}{$\times$} & \textcolor{red}{$\times$}& \textcolor{red}{$\times$} & \textcolor{green}{\checkmark} &  \textcolor{red}{$\times$} & \textcolor{red}{$\times$}  &  \textcolor{red}{$\times$}&  \textcolor{red}{$\times$}&  \textcolor{red}{$\times$}\\

      \cite{zeng2024barking} & \textcolor{red}{$\times$}&\textcolor{green}{\checkmark}& \textcolor{red}{$\times$} &\textcolor{green}{\checkmark} &\textcolor{red}{$\times$}&\textcolor{red}{$\times$}   &\textcolor{red}{$\times$} & \textcolor[gray]{0}{-} &  \textcolor[gray]{0}{-}& \textcolor[gray]{0}{-} & \textcolor[gray]{0}{-}  &  \textcolor[gray]{0}{-}&  \textcolor[gray]{0}{-}\\

      \cite{chen2024u} &\textcolor{red}{$\times$} & \textcolor{green}{\checkmark}& \textcolor{green}{\checkmark} & \textcolor{red}{$\times$} &\textcolor{red}{$\times$} &\textcolor{red}{$\times$} & \textcolor{red}{$\times$}  & \textcolor{green}{\checkmark} & \textcolor{red}{$\times$} & \textcolor{red}{$\times$}  &  \textcolor{red}{$\times$}&  \textcolor{red}{$\times$}&  \textcolor{red}{$\times$}\\
      
    \bottomrule
    \end{tabular}
    }
    \caption{Main methods for automatic and human evaluation of pun generation. PPLs., D1\&2., Succ., Ambi., Dist., Surp., and Unus. denote the metrics of Perplexity Score, Dist-1 \& Dist-2, Structure Succ., Ambiguity, Distinctiveness, Surprisal, and Unusualness, respectively. Similarly, Succ., Funn., Gram., Flun., Info., Cohe., and Read. represent Success, Funniness, Grammar, Fluency, Informativeness, Coherence, and Readability. \textcolor{green}{\checkmark} indicates metrics that are used, while \textcolor{red}{$\times$} indicates metrics that are not used. The symbol ``-'' signifies that the method is not applicable to this evaluation.}
    \label{human evaluation}
\end{table*}

In this section, we examine both automatic and human evaluation methods for pun generation. Table \ref{human evaluation} summarizes the primary metrics for evaluation and more details are provided in the Appendix \ref{additional}. 
\subsection{Automatic Evaluation}
The automatic evaluation metrics can be categorized into funniness, diversity and fluency based on the intention and definition.
\subsubsection{Funniness}
\textbf{Ambiguity \& Distinctiveness.}
\citet{kao2016computational} introduced the metrics of \textit{ambiguity} and \textit{distinctiveness} based on information theory. These metrics integrate computational models of general language understanding and pun features to quantitatively predict humour with fine-grained precision \cite{kao2016computational}. Specifically, ambiguity refers to the uncertainty arising from multiple possible meanings within a sentence, which is formulated as: 
\begin{equation}
Amb(M) = -\sum_{k \in \{a, b\}} P(m_k \mid \vec{w}) \log P(m_k \mid \vec{w})
\end{equation}
where $\vec{w}$ is a vector of observed content words in a sentence and $m_k$ is the latent sentence meaning. Higher ambiguity allows the sentence to better support both the pun and its alternative meanings.

Distinctiveness evaluates the differences between word sets that support distinct meanings within a sentence using the symmetrized Kullback-Leibler divergence $D_{KL}$, defined as follows:
\begin{equation}
Dist(F_a, F_b) = D_{KL}(F_a \| F_b) + D_{KL}(F_b \| F_a)
\end{equation}

where $F_a$ and $F_b$ represent the set of words in a sentence that support two different meanings along with their probability distributions. The high distinctiveness indicates that the distributions of the two-word groups differ significantly, which enhances the humourous effect.


\textbf{Surprisal.} Surprisal is a quantitative metric for surprise based on the pun word and the alternative word given local and global contexts \cite{he2019pun}. The formulation of local surprisal and global surprisal are defined as follows:
\begin{equation}
\begin{aligned}
&S_{\text{local}} = \coloneqq S(x_{p-d:p-1}, x_{p+1:p+d}), \\
&S_{\text{global}} = \coloneqq S(x_{1:p-1}, x_{p+1:n}),
\end{aligned}
\end{equation}
where $S$ is the log-likelihood ratio of two events, $x_1, \dots, x_n$ is a sequence of tokens, $p$ is the pun word and $d$ is the local window size. Finally, a unified metric is defined as a ratio of local-global surprisal to quantify the success of pun generation.



\subsubsection{Diversity}

\textbf{Unusualness.}
Given the uniqueness of puns, \textit{unusualness} measures based on the normalised log probabilities from language models are also utilised for pun evaluation \cite{he2019pun,pauls2012large}, which is formulated as follows: 
\begin{equation}
Unusualness \overset{def}{=} -\frac{1}{n} \log \left( \frac{p(x_1, \dots, x_n)}{\prod_{i=1}^n p(x_i)} \right)
\end{equation}
where $p(x_1, \dots, x_n)$ and $p(x_i)$ are the joint and independent probabilities, respectively. A higher metric result suggests the presence of uncommon collocations, innovative sentence structures, and other linguistic features, aligning with the characteristics of puns.

\textbf{Dist-1 \& Dist-2.} Dist-1 and Dist-2 focus on the diversity of words and phrases in the generated text \cite{li2015diversity}, which calculates the proportion of unique n-grams to the total number of n-grams, as formulated Dist-1, for example: 
\begin{equation}
\text{Dist-1} = \frac{unique \; unigrams}{total\; generated\; words}
\end{equation}
\begin{equation}
\text{Dist-2} = \frac{unique \; bigrams}{total \; generated \; bigrams}
\end{equation}
where a higher Dist-1 and Dist-2 score indicates greater diversity in the generated sentences, whereas a lower score suggests more generic and repetitive text.

\subsubsection{Fluency}
\textbf{Perplexity score} \cite{jelinek1977perplexity}. This score evaluates whether the generated puns are natural and fluent. In practice, some studies \cite{yu2018neural} quantified by using the generative language model, formally described as follows:
\begin{equation}
perplexity = \exp\left( - \frac{1}{N} \sum_{i=1}^{N} \log P(x_i|x_{<i}) \right)
\end{equation}
where $P(x_i|x_{<i})$ is the probability of the $i$-th token of a pun, given the sequence of tokens ahead.

\textbf{Structure Succ.} The evaluation measures the rate of contextual word and pun word integration, specifically the proportion of successful inclusion of pun words in the generated puns, formally shown as follows:
\begin{equation}
Succ = \frac{t_{correct}}{T} \times 100\%
\end{equation}
where $t_{correct}$ is the number of generated puns with correctly included pun words and $T$ is the total number of generated puns.

\subsection{Human Evaluation}
In the task of pun generation, since puns are a creative form of language \cite{yu-etal-2020-homophonic}, human evaluation is essential and intuitively assesses the quality of the generated puns. The primary evaluation metrics are: \textbf{Success} recognises whether the generated sentence qualifies as a successful pun based on the definition from \cite{miller2017semeval}; \textbf{Funniness} evaluates the humour and comedic quality of the generated sentences; \textbf{Fluency} shows whether the sentence is grammatically correct and flows naturally; \textbf{Informativeness} rates whether the generated sentences effectively convey meaningful and specific information; \textbf{Coherence} assesses the logical consistency and contextual suitability of word senses in the generated sentence; \textbf{Readability} indicates whether the sentence is easy to understand semantically.

Most studies utilize the Likert Scale \cite{likert1932technique} to assess the metrics. This commonly used psychological measurement method and relies on numerical scales within a specific range to evaluate a given objective \cite{Alhazmi2024DistractorGI}. For example, \citet{mittal2022ambipun} utilized a Likert scale ranging from 1 (not at all) to 5 (extremely) to rate the funniness and coherence of puns. In particular, for success metrics, some studies adopt a binary classification method in which evaluators determine whether the generated pun is successful by selecting \textit{True} or \textit{False} \cite{,tian2022unified,sun2022context,chen2024u}.

With the development of LLMs, \citet{chen2024u} conducted a human A/B test, asking annotators to compare paired puns generated by their methods and ChatGPT and select more humorous puns. Since GPT-4's evaluations aligned closely with those of human reviewers \cite{liang2024can}, \citet{zeng2024barking} replaced human reviewers with GPT-4 to assess the metrics of readability, funniness, and coherence.


\section{Challenges and Future Directions}
\label{future}
This section outlines the challenges and explores potential directions for future work.

\subsection{Multilingual Research}
With advancements in pun generation research, the majority of studies focus primarily on English, as shown in Table \ref{methods1}, while studies on puns in other languages remain limited. Linguistically, different languages employ distinct mechanisms to create puns. For example, ideographic or mixed languages, such as Chinese and Japanese,tend to construct puns across multiple linguistic and cultural levels \cite{shao2013contrastive}, such as pictographic form.  More details of linguistics in other languages are provided in the Appendix \ref{Multilingual Puns}. Therefore, cross-language pun generation can also serve as a potential future work. Building on previous cross-linguistic research, using parallel data, including word-parallel \cite{zhao2020inducing,alqahtani2021using} and sentence-parallel \cite{reimers2020making,heffernan2022bitext}, can be utilized to achieve targeted alignment of pun words.  Additionally, some pioneering works can capture phonological and semantic puns through advanced learning approaches such as contrastive learning \cite{hu2024comprehensive}, modify pre-training schemes \cite{clark2020electra} and adapter tuning \cite{parovic2022bad}.


\subsection{Multi-Modal Information}
Multimodal information enables a more reliable understanding of the world \cite{stein1993merging}, and incorporating multiple modalities into tasks can enhance the quality of pun generation. Although previous studies have introduced some multimodal evaluations and datasets \cite{zhang2024creating,chung2024can}, few have specifically focused on the generation of multimodal puns. One potential method is shared representation \cite{ngiam2011multimodal}, which involves integrating complementary information from different modalities to learn higher-performance representations \cite{lahat2015multimodal}. For example, automatic speech recognition \cite{malik2021automatic} can be leveraged to enhance homophonic puns. Another direction is to translate puns between modalities, i.e., cross-modal generation \cite{suzuki2022survey}, including text-to-image \cite{zhang2023text}, image-to-text \cite{he2017deep}, text-to-speech \cite{zhang2023survey} and speech-to-text \cite{fortuna2018survey}


\subsection{PLMs Prompting Design}
While prompt engineering has proven effective in enhancing text generation capabilities of LLMs \cite{liu2023pre}, current research still faces significant limitations in generating puns, such as an over-reliance on overly simplistic or single-faceted prompts. Chain-of-thought prompting is a powerful technique that significantly improves the reasoning capabilities of LLMs \cite{wei2022chain}. Therefore, the quality of pun generation can be enhanced by transferring CoT technique from other fields, such as using iterative bootstrapping \cite{sun2023enhancing}, knowledge enhancement \cite{dhuliawala2023chain,he2024exploring}, question decomposition \cite{trivedi2022interleaving} and self-ensemble \cite{yin2024aggregation}. Furthermore, the resut can be improved by optimizing CoT's prompt construction, including by semi-automatic prompting \cite{shum2023automatic} and automatic prompting \cite{zhang2022automatic}, as well as exploring diverse topological variants \cite{chu2024navigate}, such as chain structures \cite{olausson2023linc}, tree structures \cite{ning2023skeleton}, and graph structures \cite{besta2024graph}.


\section{Conclusion}
\label{conclusion}
In this paper, we present a comprehensive survey on pun generation tasks, including phonetic, graphic and visual puns. We classify and thoughly analyse the datasets used in pun research, review previous approaches to pun generation, discuss existing methods, as well as summarize the evaluation metrics for pun generation. Furthermore, we highlight the challenges and future directions, offering insights for researchers interested in pun generation. To enhance the research, we plan to provide an updated reading list available on the GitHub repository.

\section*{Limitations}

Although we have attempted to extensively analyse the existing literature on pun generation, some works may still be missed due to variations in search keywords. Furthermore, our exploration of other categories of puns is limited, such as recursive puns and antanaclasis, as we encountered challenges while searching for them, which may be influenced by the relatively low attention they have received in the research community. Finally, due to the rapid development of the research field, this study does not cover the entire historical scope nor the latest advancements following the survey. However, our work represents the first comprehensive survey on pun generation, including datasets, methods, evaluation, challenges and potential directions, making it a valuable resource for scholars in this field.

\section*{Acknowledgments}
This research is supported by the Strong AI Lab and the Natural, Artificial, and Organisation Intelligence Institute at the University of Auckland. The first author of this research is funded by the China Scholarship Council (CSC). 

\bibliography{custom}

\appendix

\section{Pun Categories}
\label{puncategories}
We outline the characteristics of different types of puns for clearer differentiation, including phonetic, graphic, meaning, and example, as shown in Table \ref{pun categories}. "Same", "similar" and "different" respectively indicate whether the pun word and its substitute word same, similar, or different in phonic, graphic and meaning.

\begin{table*}
    \centering
    \renewcommand{\arraystretch}{1.5} 
    \resizebox{2\columnwidth}{!}{
    \begin{tabular}{c<{\centering} c<{\centering} c<{\centering} c<{\centering} m{7cm}}
    \toprule
         Type & Phonetics & Graphic & Meaning &  Example\\
    \midrule
         Homophonic Puns & Similar & Different & Different & Dentists don't like a hard day at the \underline{orifice} (office).\\
         Heterographic Puns  & Same & Different & Different & Life is a puzzle, look here for the missing \underline{peace} (piece).\\
         Homographic Puns & Same & Same & Different & Always trust a glue salesman. They tend to \underline{stick} to their word.\\
         Visual Puns & N/A & N/A & Different & \includegraphics[width=0.6\linewidth]{image.png} \\
    \bottomrule
    \end{tabular}
    }
    \caption{List of pun categories. N/A indicates that the element is not applicable.}
    \label{pun categories}
\end{table*}

\begin{table*}
  \centering
  \resizebox{2\columnwidth}{!}{
  \begin{tabular}{lccccc}
    \toprule
    \textbf{Dataset}   & \textbf{Type} & \textbf{Source} & \textbf{Corpus (C)} & \textbf{Language} & \textbf{Availability} \\
    \midrule
    Paron\cite{sobkowiak1991metaphonology} &heg&Advertisements &3,850&English&\textcolor{green}{\checkmark}\\
    Paron-edit\cite{hempelmann2003paronomasic} &heg&\cite{sobkowiak1991metaphonology}& 1,182 & English& \textcolor{red}{$\times$}\\
    Church\cite{bell2011wordplay} &hog&Church&373 & English & \textcolor{red}{$\times$} \\
    Pun-Yang\cite{yang2015humor} &N/A&Website&2,423 & English& \textcolor{green}{\checkmark}\\
    Pun-Kao\cite{kao2016computational} & hop &Website& 435 & English& \textcolor{green}{\checkmark}\\
    Puns \cite{jaech2016phonological}&N/A&Website&75&English& \textcolor{red}{$\times$}\\
    SemEval \cite{miller2017semeval} & hog\&heg & Experts &2,878 & English& \textcolor{green}{\checkmark}\\
    SemEval-P \cite{miller2017semeval} & hog & Experts &1,607 & English& \textcolor{green}{\checkmark}\\
    SemEval-G \cite{miller2017semeval} & heg & Experts & 1,271 & English& \textcolor{green}{\checkmark}\\
    
    ExPUNations \cite{sun2022expunations}  &hog\&heg&\cite{miller2017semeval}& 1,999 &English  & \textcolor{green}{\checkmark}  \\
    CUP \cite{sun2022context} &hog\&heg& \cite{miller2017semeval}&2,396 &English & \textcolor{green}{\checkmark}        \\
    ChinesePun \cite{chen2024u}  &hop\&hog&Website&2,106 & Chinese & \textcolor{green}{\checkmark} \\
    ChinesePun-P \cite{chen2024u}  &hop&Website&1,049 & Chinese & \textcolor{green}{\checkmark} \\
    ChinesePun-G \cite{chen2024u}  &hog&Website&1,057 & Chinese & \textcolor{green}{\checkmark} \\
    Pun Rebus Art \cite{zhang2024creating} &visual&Museum&1,011& Multi-language& \textcolor{green}{\checkmark} \\
    UNPIE \cite{chung2024can} & hog\&heg & \cite{miller2017semeval} & 1,000 & Multi-language& \textcolor{green}{\checkmark}\\
    UNPIE-P \cite{chung2024can} & hog & \cite{miller2017semeval} &500& Multi-language& \textcolor{green}{\checkmark}\\
    UNPIE-G \cite{chung2024can} & heg & \cite{miller2017semeval} &500& Multi-language& \textcolor{green}{\checkmark}\\

    \bottomrule
  \end{tabular}
  }
  \caption{
    List of pun datasets. Hog, hop, heg and visual denote the types of homographic puns, homophonic puns, heterographic puns and visual puns, respectively. N/A indicates that the elements are not mentioned in the original paper.
  }
  \label{dataset}
\end{table*}

\begin{table*}
    \centering
    \resizebox{1.7\columnwidth}{!}{
    \begin{tabular}{lccc}
        \toprule
        \textbf{System}  & \textbf{Type} & \textbf{Task}  & \textbf{Language}\\
         \midrule
         JAPE \cite{binsted1994implemented} & heg \& hog & Question-Answer & English \\
         HCPP \cite{venour2000computational} & hop& Text Generation & English \\
         WISCRAIC \cite{mckay2002generation} & heg & Text Generation & English \\
         PUNDA \cite{dybala2008humor} & heg \& hog & Dialogue & Japanese \\
         STANDUP \cite{manurung2008construction} &hop &Dialogue &  English \\
         T-PEG \cite{hong2009automatically} & hop \& hog & Text Generation & English \\
         PAUL BOT \cite{tyler2020computational} &hop \& hog&Dialogue& English\\
         AliGator \cite{dehouck2025rule}&hop&Text Generation& French \\
         \bottomrule
    \end{tabular}
    }
    \caption{System of pun generation using conventional methods. Hog, hop and heg denote the types of homographic puns, homophonic puns and heterographic puns, respectively.}
    \label{traditional}
\end{table*}

\section{Additional Evaluation}
\label{additional}
In this section, we outline the limitations of the evaluation metrics and supplement additional supporting details.
\subsection{Limitations}
\subsubsection{Automatic Evaluation}
Methods such as Surprisal-based evaluation are influenced by context dependency. In particular, local Surprisal is highly sensitive to the choice of the local window size. In addition, metrics such as Dist-1 and Dist-2, which measure lexical and n-gram diversity based on statistical and information-theoretic principles, fail to capture semantic diversity. Similarly, the Perplexity score (PPLs) evaluates text based on the probability of model-generated words, where a lower PPLs indicates better predictive performance but does not necessarily imply semantic coherence or logical consistency.

\subsubsection{Human Evaluation}
Although human evaluation is considered the gold standard, it still exhibits a significant degree of subjectivity in metrics such as readability and funniness. This subjectivity primarily stems from differences in participants' cultural backgrounds and knowledge levels \cite{chen2023individual}. However, many studies claim to have employed qualified workers or annotators, while they failed to provide detailed information about the evaluators' backgrounds, which can easily lead to variability in the final assessments. Therefore, imposing clearer selection criteria for participants may help mitigate the impact of subjectivity in evaluation.

\subsection{Supplement Details}
\textbf{Suprisal.} Based on \cite{he2019pun}, the pun word $w^p$ is more surprising relative to its alternative word $w^a$ in the local context, while is less in the global context. Therefore, $S_{\text{ratio}}$ is defined as a ratio to balance the metric:
\begin{equation} 
\resizebox{0.9\hsize}{!}{$
S_{ratio} \coloneqq
\begin{cases} 
-1, & S_{local} < 0 \text{ or } S_{global} < 0, \\
S_{local}/{S_{global}}, & \text{otherwise}.
\end{cases}
$}
\end{equation}
where $S_{local}$ and $S_{local}$ are local surprisal and global surprisal, respectively. A higher value of $S_{ratio}$ indicates a better-quality pun.


\section{Dataset}
\label{pundataset}
The pun dataset for different types are summarized in Table \ref{dataset}. We list the datasets in five dimensions: 
\begin{itemize}
    \item The type of puns.
    \item The source of the datasets.
    \item The total number of the datasets.
    \item The language of the datastes.
    \item Is the dataset publicly available?
\end{itemize}
Early pun datasets, such as Paron \cite{sobkowiak1991metaphonology} and Church \cite{bell2011wordplay}, were primarily constructed from publicly available sources with a strong preference for specific domains, such as advertisements, church and newspaper comics, which are more suitable for use in domain-specific applications. Among the listed datasets, SemEval \cite{miller2017semeval} is the first expert-annotated pun dataset, covering both homophonic and heterographic puns, and has since become the most widely references in subsequent research. Furthermore, recent developments have introduced some multimodal and multilingual pun datasets, which have expanded the scope and potential directions for research in pun generation.

\section{Paper Collection}
This section outlines the approach that we used to collect relevant papers in this survey. We initially searched for the keywords "pun research", "computational humour", and "pun dataset" on arXiv and Google Scholar, identifying a total of around 150 publications. Then, we filtered the papers that specifically focused on pun generation, resulting in approximately 30 papers. Subsequently, we applied the forward and backward snowball technique by examining the references and citations of these seed papers to identify additional relevant studies. We carefully reviewed all identified papers and ultimately compiled the findings into this survey.

\section{Conventional Systems}
In this section, we summarize the pun generation systems with conventional methods in Section \ref{Traditional Models}, as shown in table \ref{traditional}. We here list the types of puns, task scenarios and languages corresponding to the system's applications.


\section{Related Surveys}
To our knowledge, there are currently only surveys on computational humour research, while no focusing exclusively on puns. \citet{amin2020survey} provides a survey on humour generation, including generation systems, evaluation methods, and datasets. However, it does not specifically analyze the category of puns and only summarizes papers published prior to 2020. \citet{nijholt2017humor} concluded a survey on designing humour and interacting with social media, virtual agents, social robots and smart environments. In addition, other humour studies have been examined from the perspectives of detection \cite{ramakristanaiah2021survey,ganganwar2024sarcasm} and recognition \cite{kalloniatis2024computational}. Furthermore, there are some relevant surveys on creating writing, such as composition of poetry \cite{bena2020introducing,elzohbi2023creative}, storytelling \cite{gieseke2021survey,alhussain2021automatic}, arts \cite{shahriar2022gan} and metaphor \cite{rai2020survey,ge2023survey}. Our survey provides a comprehensive overview of various methods focused on pun generation, including those published in recent years.
\section{Potential Research in Visual Puns}
\label{potential Research}
In Section \ref{visual-language}, we reviewed studies on visual puns. However, to the best of our knowledge, research on the generation and evaluation of visual puns remains limited. Existing research primarily leverages multimodal models to generate textual descriptions incorporating visual pun elements as an intermediate task, using visual cues to aid in the comprehension of textual puns \cite{rebrii2022words,chung2024can}. Therefore, text-to-image generation presents a promising research direction in this field, as it can help mitigate comprehension challenges that arise in single-modality interpretation.

One potential approach is to simulate the multimodal training paradigm of CLIP \cite{radford2021learning} by constructing a pun-specific semantic vector space based on pun corpora. For text-to-image generation, this method would first encode the dual meanings of the pun, integrating both its original and pun-specific semantics, and then generate visual pun images by aligning them within the trained pun semantic space. For example, a mousetrap catches a white mouse, as illustrated in Figure \ref{visual pun}. The word mouse can refer to both an animal and an electronic device. By encoding the dual meanings of this sentence, the trained pun-specific semantic space can generate a corresponding visual pun representation.

Additionally, multimodal approaches may be particularly suitable for non-English languages that rely on strokes rather than spelling. For example, in Chinese, certain character errors or newly coined characters can create pun-like effects, triggering humour through visual wordplay. Finally, models such as DeepFloyd IF \cite{DeepFloydIF}, Stable Diffusion v1–5 \cite{rombach2021highresolution}, and DALL-E \cite{ramesh2022hierarchical}, which are based on variational auto-encoders, diffusion models, and autoregressive models, also offer powerful image generation capabilities. While these models are not specifically designed for visual pun generation, integrating pun-related features could make them a promising direction for this task.

\section{Application}
This section explores the relevance of pun generation within the broader field of natural language generation (NLG) and its diverse real-world applications. As a creative NLG task, pun generation leverages semantic ambiguity and phonetic similarity to produce humorous and engaging text, thereby enhancing the expressive capabilities of language models. Its applications span across advertising, conversational agents, education, and entertainment, highlighting its potential to foster user engagement and stimulate creativity in practical contexts.
\subsection{Relevance}
Pun generation is a specialized NLG task that shares core objectives with broader NLG, such as generating coherent and contextually appropriate text~\cite{gatt2018survey}. However, its focus on humour and wordplay introduces unique challenges,  requiring models to balance polysemy, phonetics, and coherence.  Methodologies like Sequence-to-Sequence models and fine-tuned pre-trained language models (PLMs), as used in ~\cite{yu2018neural} for puns and ~\cite{raffel2020exploring} for NLG tasks, highlight shared technical foundations. Pun generation advances NLG by improving models’ handling of semantic ambiguity, as seen in ~\cite{luo2019pun}, which introduced ambiguity rewards.  Recent prompting strategies, such as those in ~\cite{xu2024good}, enhance NLG creativity, benefiting tasks like dialogue generation. By tackling these complexities, pun generation drives innovations in NLG, particularly in multilingual and multimodal contexts ~\cite{chung2024can}.

\subsection{Applications}
Pun generation finds practical utility across multiple domains. In advertising,  puns create memorable slogans, as seen in KitKat’s 2023 campaign, “Have a break, have a KitKat, "playing on break" as pause and physical snap~\cite{kitkat2023campaign}. ~\citet{puns_llms2024} showed LLMs like GPT-4 can generate coherent advertising puns, which helps marketers. In conversational systems, puns enhance engagement,  with Google Assistant using phrases like “I’m on a roll’’ for baking queries ~\cite{google2024assistant}. ~\citet{chen2024u} fine-tuned LLaMA2 for dialogue puns, improving user satisfaction. In education, puns foster linguistic creativity, as demonstrated  by ~\cite{tyler2020computational}. PAUL BOT, which aids children’s communication. In entertainment, puns enrich narratives and gaming, with ~\cite{chung2024can} using DALL-E 3 for visual puns in interactive storytelling. Future applications  include personalized marketing and therapeutic humor, leveraging multimodal  models to create immersive experiences.


\section{Multilingual Puns}
This section introduce morphological process in different languages, pun research from linguistic perspective and their linguistic resource available.
\label{Multilingual Puns}
\subsection{Morphological Process}

We outline the mainly morphological process of different languages to analyze potential approaches for multilingual pun processing. Table \ref{Morphological} shows the application of various morphological processes in English, Chinese, Arabic, Spanish, French and Japanese.


\textbf{Derivation} refers to the process of forming a new word by adding an affix (such as a prefix or suffix) to a root or stem \cite{beard2017derivation}, which is the most popular in different language. 

\textbf{Compounding} is the morphological process of creating new words by combining two or more independent words or word roots \cite{to2025definition}. This process plays a particularly important role in Chinese, where compound words are highly prevalent. As a result, the majority of Chinese characters used in word formation tend to carry dual or multiple meanings \cite{arcodia2007chinese}. 

\textbf{Clipping} is the process of whereby a multisyllabic word is shortened by removing one or more of its parts, such as back-clipping, fore-clipping and mixed clipping \cite{ishchenko2023issue} to form a new, shorter word. This morphological process is observed in several languages, including French and Japanese \cite{hamilton2024clipping}. 

\textbf{Borrowing} is the way of incorporating lexical items from other languages directly into the native lexicon \cite{haspelmath2009lexical}. It is worth noting that word formation through borrowing is particularly common in Chinese \cite{cannon1988chinese}, English and Japanese \cite{rao2018significance,stanlaw1987japanese}. For example, a large number of English words originate from Latin, French, Greek, and other languages \cite{green2020greek}, such as cliche and cuisine (from French). 

\textbf{Conversion} refers to the process of assigning a new grammatical function or part of speech to an existing word without altering its form \cite{tachmyradova2020conversion}. Compared to other languages, English has the extremely prevalent phenomenon \cite{balteiro2006contribution}. 

\textbf{Reduplication} involves the repetition of all or part of a word to convey various grammatical meanings, rhetorical effects, or expressive tones \cite{moravcsik1978reduplicative}, including Chinese \cite{xu2012reduplication}, French \cite{scullen2008new} and Japanese \cite{balouglu2022category}. 

\textbf{Onomatopoeia} refers to the formation of words that phonetically imitate the sounds associated with natural phenomena or actions \cite{bredin1996onomatopoeia}. Some studies focus on languages characterized by lexicons rich in sound-symbolic expressions, especially in African and Asian languages such as Japanese \cite{ohala1997sound}.

Understanding morphological process can provide valuable insights into the mechanisms underlying pun generation. For example, conversion shows some certain similarities with homographic puns, as both involve assigning different meanings or grammatical functions to the same spelling. Therefore, examining the morphological strategies that are prevalent in different languages provide a promising direction for exploring multilingual pun generation.
\begin{table}
    \centering
    \resizebox{0.9\columnwidth}{!}{
    \begin{tabular}{ccccccc}
        \toprule
        \textbf{MoP} &\textbf{En.}  & \textbf{Ch.} & \textbf{Ar.} & \textbf{Sp.}& \textbf{Fr.}& \textbf{Ja.}\\
         \midrule
         Derivation  & \greenTriangle & \greenTriangle & \greenTriangle  & \greenTriangle & \greenTriangle & \greenTriangle\\
         Compounding   & \greenTriangle  & \greenTriangle & \redCircle & \redCircle & \redCircle & \greenTriangle\\
         Clipping & \greenTriangle & \redCircle &  \redCircle&  \redCircle& \greenTriangle & \greenTriangle \\
         Borrowing & \greenTriangle  &  \redCircle&  \redCircle& \redCircle & \redCircle & \greenTriangle\\
         Conversion & \greenTriangle& \redCircle&  \redCircle&  \redCircle&  \redCircle&  \redCircle\\
         Reduplication & \redCircle & \greenTriangle & \redCircle & \redCircle & \greenTriangle & \greenTriangle  \\
         Onomatopoeia & \redCircle  & \greenTriangle &\redCircle &\redCircle &\redCircle &\greenTriangle \\
         \bottomrule
    \end{tabular}
    }
    \caption{Language family characteristics and pun findings in some major languages. MoP represents the morphological process. \protect\greenTriangle ~ indicates that the morphological process is highly productive in the given language, whereas ~ \protect\redCircle ~ signifies the specific morphological process used in a limited or less research. En., Ch., Ar., Sp., Fr. and Ja. are English, Chinese, Arabic, Spanish, French and Japanese, separately.}
    \label{Morphological}
\end{table}


\subsection{Puns in Different Languages}
From a linguistic perspective, we explore some methods used for generating puns across different languages, providing insights for automatic pun generation. 

\textbf{Chinese.} Since Chinese only has about 1,300 different syllables \cite{duanmu2007phonology}, there are a large number of homophones in Chinese. This feature has enriched the forms of puns based mainly on homophones, while it has also increased the difficulty of analyzing homophonic puns. In addition, in research on logographic languages, \citet{zheng2020effect} employed the direct access model and graded salience hypothesis \cite{glucksberg1986context,giora2003our,shen2016processing} to investigate the cognitive processing of Chinese puns.

\textbf{French.} According to the CLEARPOND \cite{marian2012clearpond}. \citet{largy1996homophone} provides evidence the homophone effect can be manifested itself through the occurrence of noun-verb inflection errors. Furthermore, \citet{kerleroux2017derivationally} argue that homophony phenomena in French are primarily based on non-affixal derivational morphology, specifically conversion processes. In addition, \citet{article} found a significant relationship between word length and polysemy in French, showing that shorter words tend to have a greater number of meanings. This observation may offer useful insights for research on pun recognition and generation in French.


\textbf{Arabic.} Most current research on Arabic puns focuses on translation tasks, especially on a few Arabic anthologies. \citet{aqad2019english} investigate the semantic dimensions of puns in the translation of the Quran. \citet{mehawesh2023challenges} highlight that the Arabic root-based morphological system differs fundamentally from that of English, and that Arabic frequently employs rhythm, repetition, and syllabic patterns to enhance punning effects, while English lacks a directly comparable rhythmic system.

\textbf{Japanese.} There are some studies on Japanese puns focusing on phonological features. \citet{kawahara2009role} showed that Japanese puns need to maintain consonant similarity when they are created, and that the criterion for this depends on psychoacoustic information, while \citet{yokogawa2001generation} further quantified phonological similarity using features such as manner and place of articulation. Notably, \citet{takanashi2007orthographic} shows that using kanji and kana orthography to process Kyoka, which is a genre of playful Japanese poetry to characteristically employ puns for humour.

\textbf{Spanish.} Some studies on puns explored the pun translation in Spanish film titles into English \cite{diaz2014relevance,diaz2008worldplay}, while other studies analyzed the lexico-semantic applied in Spanish humour including homonymy, polysemy and intra-phrasal syllables \cite{bobchynets2022lexico}.



\subsection{Resource Available}
We investigate the available linguistic resources across multiple languages provide a reference on multilingual puns for future research.

\textbf{English.} There is a large corpus of material available for the study of English puns, as introduced in Section~\ref{pundata}. \textbf{Chinese.} In addition to the Chinese pun database mentioned in Section~\ref{pundata}, several open Chinese linguistic resources are also available, such as THULAC \cite{li2009punctuation}, Peking University CCL Corpus \footnote{http://ccl.pku.edu.cn:8080/ccl\_corpus/}, and CLUECorpus2020 \cite{xu2020cluecorpus2020}. \textbf{French.} Various French language resources have been developed for language modeling, including those by \cite{sagot2010lefff} and \cite{abeille2003building}. \textbf{Arabic.} \citet{jakubivcek2013tenten} constructed a large-scale Arabic general corpus using web crawling techniques. Additionally, Linguistic Data Consortium (LDC) produced Arabic Gigaword \footnote{https://catalog.ldc.upenn.edu/LDC2003T12}, which contains approximately 1 million news documents totaling 400 million words of Arabic text. \textbf{Spanish.} Some Spanish linguistic resources have been developed by Real Academia Española (RAE) such as CREA \cite{sanchez2005corpus} and \cite{rojo2016corpes} which provide extensive collections of both written and spoken samples from Latin American and European varieties of Spanish. \textbf{Japanese.} A range of Japanese corpora are available for lexicological research and language modeling. Notable examples include the Balanced Corpus of Contemporary Written Japanese (BCCWJ) \cite{maekawa2014balanced}, the Corpus of Spontaneous Japanese (CSJ) \cite{maekawa2000spontaneous} and jaTenTen, a web corpus compiled for large-scale linguistic analysis \cite{jakubivcek2013tenten}.

\section{Puns in LLMs}
Puns are considered a valuable tool for evaluating LLMs in their ability to understand linguistic humour and wordplay \cite{xu2024good}. They help reveal the models’ capabilities and limitations in tasks that require semantic ambiguity, phonetic similarity, and contextual reasoning. Specifically, puns enable a systematic assessment of LLMs' proficiency in nuanced linguistic reasoning within creative language applications, particularly in tasks such as pun recognition, explanation, and generation~\cite{large2019,dsilva2024}.

Recent studies~\cite{puns_llms2024,aihumorgen2025} have revealed several insights regarding puns in LLMs: (1) While most large language models (LLMs) are highly sensitive to prompt bias in recognition tasks, some demonstrate more stable performance and achieve higher recognition accuracy. Moreover, their performance can be further improved by incorporating definitions and examples. (2) Most LLMs are capable of recognizing pun words. Although alternative words may not significantly affect the recognition of a pun, they play an important role in clearly explaining its meaning. Some LLMs demonstrate explanation quality comparable to, or even surpassing, that of humans. However, common errors observed among LLMs include: incorrect identification of pun type, misidentification of the pun word and insufficient analysis of the dual meanings. (3) LLMs show particular skill in generating homographic puns. Providing contextual words significantly improve the quality of these puns. However, a ``Lazy Pun Generation'' pattern has been observed, where the model tends to reuse the same pun words repeatedly, indicating a lack of creativity. While some of LLMs have achieved state-of-the-art performance in generation tasks, their humour generation still falls short compared to that of humans.

\end{document}